\def\year{2020}\relax
\documentclass[letterpaper,draft]{article} 
\usepackage{aaai20}  
\usepackage{times}  
\usepackage{helvet} 
\usepackage{courier}  
\usepackage[hyphens]{url}  
\usepackage{graphicx} 
\usepackage{graphicx}  
\def\sectionflex#1{\subsection{#1}}

\usepackage{pgfplots}
\usepackage{amsmath}
\usepackage{amssymb}
\usepackage{tikz}
\usepackage{booktabs}
\usepackage{arydshln}
\usepackage{tabularx}
\usepackage{multirow}
\usepackage{filecontents}

\title{Rare Words: A Major Problem for Contextualized Embeddings \\ 
and How to Fix it by Attentive Mimicking}
\author{Timo Schick \\ Sulzer GmbH \\ Munich, Germany \\ \url{timo.schick@sulzer.de} \And Hinrich Sch\"utze 
\\ Center for Information and Language Processing \\ LMU Munich, Germany \\ \url{inquiries@cislmu.org}}

\pgfplotsset{compat=1.13}
\newcommand\keyword{\texttt{<W>}}
\newcommand\dataset{WNLaMPro}
\newcommand\mask{\underline{\hspace{0.3cm}}}
\newcommand{\bertbase}{\ensuremath{\text{BERT}_\textsc{base}}}
\newcommand{\bertlarge}{\ensuremath{\text{BERT}_\textsc{large}}}

\makeatletter
\def\adl@drawiv#1#2#3{%
        \hskip.5\tabcolsep
        \xleaders#3{#2.5\@tempdimb #1{1}#2.5\@tempdimb}%
                #2\z@ plus1fil minus1fil\relax
        \hskip.5\tabcolsep}
\newcommand{\cdashlinelr}[1]{%
  \noalign{\vskip\aboverulesep
           \global\let\@dashdrawstore\adl@draw
           \global\let\adl@draw\adl@drawiv}
  \cdashline{#1}
  \noalign{\global\let\adl@draw\@dashdrawstore
           \vskip\belowrulesep}}
\makeatother

\newcounter{notecounter}

\newcommand{\enoteson}{\long\gdef\enote##1##2{{
\stepcounter{notecounter}
{\large\bf
\hspace{1cm}\arabic{notecounter} $<<<$ ##1: ##2
$>>>$\hspace{1cm}}}}}
\enoteson

\def\secref#1{\S\ref{sec:#1}}
\def\seclabel#1{\label{sec:#1}}
\def\eqref#1{Eq.~\ref{eqn:#1}}

\newcommand\namecite[1]{\citeauthor{#1} (\citeyear{#1})}
\newcommand{\citet}[1]{\citeauthor{#1} \shortcite{#1}}
\newcommand\citenp[1]{\citeauthor{#1} \citeyear{#1}}

\definecolor{col1}{cmyk}{1,0.3968,0,0.2588} 
\definecolor{col2}{cmyk}{0,0.6175,0.8848,0.1490} 
\definecolor{col3}{cmyk}{0.1127,0.6690,0,0.4431} 
\definecolor{col4}{cmyk}{0.6765,0.2017,0,0.0667} 
\definecolor{col5}{cmyk}{0.3081,0,0.7209,0.3255} 
\definecolor{col6}{cmyk}{0,0.8765,0.7099,0.3647}

\begin{document}

\maketitle

\begin{abstract}
Pretraining deep neural network architectures with a language modeling objective has brought large improvements for many natural language processing tasks. Exemplified by BERT, a recently proposed such architecture, we demonstrate that despite being trained on huge amounts of data, deep language models still struggle to understand rare words. To fix this problem, we adapt Attentive Mimicking, a method that was designed to explicitly learn embeddings for rare words, to deep language models. In order to make this possible, we introduce \emph{one-token approximation}, a procedure that enables us to use Attentive Mimicking even when the underlying language model uses subword-based tokenization, i.e., it does not assign embeddings to all words.
To evaluate our method, we create a novel dataset that tests the ability of language models to capture semantic properties of words without any task-specific fine-tuning. Using this dataset, we show that adding our adapted version of Attentive Mimicking to BERT does substantially improve its understanding of rare words. 
\end{abstract}
\section{Introduction}

Distributed representations of words are a key component of
natural language processing (NLP) systems. In particular,
deep contextualized representations learned using an
unsupervised language modeling
objective \cite{peters2018deep} have led to large
performance gains for a variety of NLP tasks.
Recently, several authors have proposed to not only use language modeling for feature extraction, but to fine-tune entire language models for specific tasks \cite{radford2018improving,howard2018universal}. Taking up this idea, \namecite{devlin2018bert} introduced BERT, a bidirectional language model based on the Transformer \cite{Vaswani2017} that has achieved a new state-of-the-art for several NLP tasks.

As demonstrated by \namecite{radford2018language}, it is
possible for language models to solve a diverse set of tasks
to some extent \emph{without} any form of task-specific
fine-tuning. This can be achieved by simply presenting the
tasks in form of natural language sentences that are to be
completed by the model. The very same idea can also be used
to test how well a language model understands a given word:
we can ``ask'' it for properties of that word using natural
language. For example, a language model that understands the
concept of ``guilt'' should be able to correctly complete
the sentence ``Guilt is the opposite of \mask.'' with the word ``innocence''. 

The examples in Table~\ref{motivational-examples} show that, according to this measure, BERT is indeed able to understand frequent words such as ``lime'' and ``bicycle'': it predicts, among others, that the former is a fruit and the latter is the same as a bike. However, it fails terribly for both ``kumquat'' and ``unicycle'', two less frequent words from the same domains. This poor performance raises the question whether deep language models generally struggle to understand rare words and, if so, how this weakness can be overcome.

\begin{table}
\begin{tabularx}{\linewidth}{lX}
\toprule
\textbf{Q:} A \emph{lime} is a \mask{} . &
\textbf{A:} lime, lemon, fruit \\[0.1cm]
\textbf{Q:} A \emph{bicycle} is a \mask{} . &
\textbf{A:} bicycle, motorcycle, bike  \\
\midrule
\textbf{Q:} A \emph{kumquat} is a \mask{} . &
\textbf{A:} noun, horse, dog \\[0.1cm]
\textbf{Q:} A \emph{unicycle} is a \mask{} . &
\textbf{A:} structure, unit, chain \\
\bottomrule
\end{tabularx}
\caption{Example queries and most probable outputs of BERT for frequent (top) and rare words (bottom)}
\label{motivational-examples}
\end{table}

To answer this question, we create a novel dataset containing queries like the ones shown in Table~\ref{motivational-examples}. This dataset consists of (i) natural language patterns such as
\[
\text{\keyword{} is a \mask{} .}
\]
where \keyword{} is a placeholder for a word to be
investigated, and (ii) corresponding pairs of keywords (\keyword{}) and
targets (fillers for \mask{}) obtained using semantic relations extracted from
WordNet \cite{miller1995wordnet}.

Using this dataset, we show that BERT indeed fails to
understand many rare words. To overcome this limitation, we
propose to apply Attentive Mimicking
\cite{schick2019attentive}, a method that allows us to
explicitly learn high-quality representations for rare
words. A prerequisite for using this method is to have
high-quality embeddings for as many words as possible,
because it is trained to reproduce known word
embeddings. However, many deep language models including
BERT make use of byte-pair encoding \cite{SennrichHB15}, WordPiece \cite{Wu2016} or
similar subword tokenization algorithms. Thus, many words
are not represented by a single token but by a sequence of
subword tokens and do not have their own embeddings.

To solve this problem, we introduce \emph{one-token
  approximation} (OTA), a method that 
approximately infers what the embedding of an arbitrary word
would look like if it were represented by a single
token. While we apply this method only to BERT, it can
easily be adapted for other language modeling architectures.

In summary, our contributions are as follows:

\begin{itemize}
\item We introduce \emph{WordNet Language Model Probing}
  (\dataset), a novel dataset
for evaluating
the ability of language models to understand specific words.
\item Using this dataset, we show that the ability of BERT to understand words depends highly on their frequency.
\item We present one-token approximation (OTA), a method
  that obtains an embedding for a multi-token word that has
  behavior similar to the sequence of its subword embeddings.
\item We apply OTA and Attentive Mimicking
  \cite{schick2019attentive} to BERT
  and show that this substantially improves BERT's understanding of rare words. \\ Our work is the first to successfully apply mimicking techniques to contextualized word embeddings.
\end{itemize}

\section{Related Work}

Using language modeling as a task to obtain contextualized representations of words was first proposed by \namecite{peters2018deep}, who train a bidirectional LSTM \cite{hochreiter1997long} language model for this task and then feed the so-obtained embeddings into task-specific architectures. Several authors extend this idea by transferring not only word embeddings, but entire language modeling architectures to specific tasks \cite{radford2018improving,howard2018universal,devlin2018bert}. Whereas the GPT model proposed by \namecite{radford2018improving} is strictly unidirectional (i.e., it looks only at the left context to predict the next word) and the ULMFiT method of \namecite{howard2018universal} uses a shallow concatenation of two unidirectional models, \namecite{devlin2018bert} design BERT as a deep bidirectional model using a Transformer architecture and a masked language modeling task.

There are roughly two types of
approaches for explicitly learning high-quality embeddings
of rare words:
surface-form-based approaches and context-based
approaches. The former use subword information to infer a
word's meaning; this includes  $n$-grams
\cite{wieting2016charagram,bojanowski2016enriching,salle2018incorporating},
morphemes \cite{lazaridou2013compositional,luong2013better}
and characters \cite{pinter2017mimicking}. On the other
hand, context-based approaches take a look at the words
surrounding a given rare word to obtain a representation for
it
(e.g., \citenp{herbelot2017high}; \citenp{khodak2018carte}). Recently,
\namecite{schick2018learning} introduced the
\emph{form-context model}, combining both approaches
by jointly using surface-form and context
information. The form-context model and its \emph{Attentive
  Mimicking} variant \cite{schick2019attentive} achieve a
new state-of-the-art for high-quality
representations of rare words.

Presenting tasks in the form of natural language sentences
was recently proposed by \namecite{mccann2018natural} as
part of their \emph{Natural Language Decathlon}, for which
they frame ten different tasks as pairs of natural language
questions and answers. They
train models on triples of questions, contexts
and answers in a supervised
fashion.
An alternative, completely unsupervised approach proposed by
\namecite{radford2018language} is  to train  a language
model on a large corpus, present text specialized
for a particular task and then let the model complete this text.
They achieve good performance
on tasks such as reading comprehension, machine translation
and question answering -- without any form of task-specific
fine-tuning.
We use this paradigm for constructing \dataset.

Several existing datasets were designed to analyze the
ability of word embeddings to capture semantic relations
between words. For example, \namecite{baroni2011we}
compile the BLESS dataset that covers five different semantic
relations (e.g., hyponymy)
from multiple sources. \namecite{weeds2014learning} also
create a dataset for semantic relations based on hypernyms
and hyponyms using WordNet
\cite{miller1995wordnet}. However, these datasets differ from
\dataset\ in two important respects. (i) They focus on frequent
words by filtering out infrequent ones whereas we
explicitly want to analyze rare words. (ii) They do not
provide natural language patterns: they either directly
evaluate (uncontextualized) word embeddings using a
similarity measure such as cosine distance or they frame the task
of identifying the relationship between two words as a
supervised task.

\section{Attentive Mimicking}

\subsection{Original Model}
\emph{Attentive Mimicking} (AM)
\cite{schick2019attentive} is a method that, given a set of
$d$-dimensional high-quality embeddings for frequent words,
can be used to infer embeddings for infrequent words that
are appropriate for the given embedding space. AM is an extension of the \emph{form-context model} \cite{schick2018learning}.

The key idea of the form-context model is to compute two distinct embeddings per word, where the first one exclusively uses the word's surface-form and the other the word's contexts, i.e., sentences in which the word was observed. Given a word $w$ and a set of contexts $\mathcal{C}$, the surface-form embedding $v_{(w,\mathcal{C})}^\text{form} \in \mathbb{R}^d$ is obtained by averaging over learned embeddings of all $n$-grams in $w$; the context embedding $v_{(w,\mathcal{C})}^\text{context} \in \mathbb{R}^d$ is the average over the known embeddings of all context words. 

The final representation $v_{(w, \mathcal{C})}$ of $w$ is
then
a weighted sum 
of form embeddings and transformed context embeddings:
\[
v_{(w,\mathcal{C})} = \alpha \cdot A v_{(w,\mathcal{C})}^\text{context} + (1 - \alpha) \cdot v_{(w,\mathcal{C})}^\text{form}
\]
where $A$ is a $d\times d$ matrix and $\alpha$ is a function
of both embeddings, allowing the model to decide when
to rely on the word's surface form and when 
on its contexts (see \namecite{schick2018learning} for further details). 


While the form-context model treats all contexts equally, AM extends it with a self-attention mechanism that is applied to all contexts, allowing the model to distinguish informative from uninformative contexts. The attention weight of each context is determined based on the idea that given a word $w$, two informative contexts $C_1$ and $C_2$ (i.e., contexts from which the meaning of $w$ can be inferred) resemble each other more than two randomly chosen contexts in which $w$ occurs. In other words, if many contexts for a word $w$ are similar to each other, then it is reasonable to assume that they are more informative with respect to $w$ than other contexts. \namecite{schick2019attentive} define the similarity between two contexts as
\[
s(C_1, C_2) =
\frac{(Mv_{C_1}) \cdot
(Mv_{C_2})^\top}{\sqrt{d}} %
\]
with $M \in \mathbb{R}^{d \times d}$ a learnable parameter
and $v_C$ denotes the average of embeddings for all words
in a context $C$. The weight of a context is then defined as
\[
\rho(C) \propto \sum_{C' \in \mathcal{C}} s(C, C')\,.
\]
with $\sum_{C \in \mathcal{C}} \rho(C) = 1$. This results in the final context embedding
\[
v_{(w, \mathcal{C})}^\text{context} = %
\  \sum_{C \in \mathcal{C}} \rho(C)\cdot v_{C} 
\] 
where again, $v_C$ denotes the average of the embeddings of all words in a context $C$.

Similar to earlier models (e.g., \citenp{pinter2017mimicking}), the model is trained through \emph{mimicking}. That is, we randomly sample words $w$ and corresponding contexts $\mathcal{C}$ from a large corpus and, given $w$ and $\mathcal{C}$, ask the model to mimic the original embedding of $w$, i.e., to minimize the squared Euclidean distance between the original embedding and $v_{(w,\mathcal{C})}$.

\subsection{AM+\textsc{context}}
\seclabel{amcontext}
As we found in preliminary experiments that AM focuses heavily on the word's surface form -- an observation that is in line with results reported by \namecite{schick2018learning} --, in addition to the default AM configuration of \namecite{schick2019attentive}, we investigate another configuration AM+\textsc{context}, which pushes the model to put more emphasis on a word's contexts.
This is achieved by (i) increasing the minimum number of sampled contexts for each training instance from $1$ to $8$ and (ii) introducing $n$-gram dropout: during training, we randomly remove $10\%$ of all surface-form $n$-grams for each training instance.

\section{One-Token Approximation}

As AM is trained through mimicking, it must be given
high-quality embeddings of many words to learn how to make
appropriate use of form and context
information. Unfortunately, as many deep language models
make use of subword-based tokenization, they assign
embeddings to comparably few words. To overcome this
limitation, we introduce \emph{one-token approximation}
(OTA). OTA finds an
embedding for a multi-token word or phrase $w$ that is similar
to the embedding that $w$ would have received if it had been
a single token. This allows us to train AM in the usual way
by simply mimicking the OTA-based embeddings of multi-token words.

Let $\Sigma$ denote the set of all characters and
$\mathcal{T} \subset \Sigma^*$  the set of all tokens
used by the language model. Furthermore, let ${t: \Sigma^*
  \rightarrow \mathcal{T}^*}$ be the tokenization function
that splits each word into a sequence of tokens and  $e:
\mathcal{T} \rightarrow \mathbb{R}^d$ the model's token
embedding function, which we extend to sequences of tokens
in the natural way as $e([t_1, \ldots, t_n]) = [e(t_1),
  \ldots, e(t_n)]$.

We assume that the language model internally consists of $l_\text{max}$ hidden layers and given a sequence of token embeddings $\boldsymbol{e} = [e_1, \ldots, e_n]$, we denote by $h_i^l(\boldsymbol{e})$ the contextualized representation of the $i$-th input embedding $e_i$ at layer $l$. Given two additional sequences of left and right embeddings $\boldsymbol{\ell}$ and $\boldsymbol{r}$, we define
\begin{align*}
\tilde{h}_i^l(\boldsymbol{\ell}, \boldsymbol{e}, \boldsymbol{r}) = 
	\begin{cases}
		h_i^l(\boldsymbol{\ell}; \boldsymbol{e}; \boldsymbol{r}) & \text{if } i \leq |\boldsymbol{\ell}| \\
		h_{i + |\boldsymbol{e}|}^l(\boldsymbol{\ell}; \boldsymbol{e}; \boldsymbol{r}) & \text{if } i > |\boldsymbol{\ell}| \\
	\end{cases}
\end{align*}
where $a ; b$ denotes the concatenation of sequences $a$ and $b$.
That is,
we ``cut out'' the sequence $\boldsymbol{e}$ and 
$\tilde{h}_i^l(\boldsymbol{\ell}, \boldsymbol{e},
\boldsymbol{r})$ is then the embedding of the $i$-th input at
layer $l$, either from $\boldsymbol{\ell}$ (if position $i$
is before $\boldsymbol{e}$) or from $\boldsymbol{r}$ (if
position $i$ is after $\boldsymbol{e}$).


To obtain an OTA embedding for an arbitrary word $w \in \Sigma^*$, we require a set of left and right contexts $\mathcal{C} \subset \mathcal{T}^* \times \mathcal{T}^*$. Given one such context $c = (\mathbf{t}_\ell,\mathbf{t}_r)$, the key idea of OTA is to search for the embedding $v \in \mathbb{R}^d$ whose influence on the contextualized representations of $\mathbf{t}_\ell$ and $\mathbf{t}_r$ is as similar as possible to the influence of $w$'s original, multi-token representation on both sequences. That is, when we apply the language model to the sequences ${s_1 = [e(\mathbf{t}_\ell); e(t(w)); e(\mathbf{t}_r)]}$ and ${s_2 = [e(\mathbf{t}_\ell); [v]; e(\mathbf{t}_r)]}$, we want the contextualized representations of $\mathbf{t}_\ell$ and $\mathbf{t}_r$ in $s_1$ to be as similar as possible to those in $s_2$. 

Formally, we define the \emph{one-token approximation} of $w$ as
\begin{align*}
& \text{OTA}({w}) = \\
& \ \arg\min_{v  \in \mathbb{R}^n} \sum_{(\mathbf{t}_\ell, \mathbf{t}_r) \in \mathcal{C}} d(e(t(w)),[v] \mid e(\mathbf{t}_\ell), e(\mathbf{t}_r))
\end{align*}
where
\begin{align*} 
d(\boldsymbol{e}, \tilde{\boldsymbol{e}} \mid \boldsymbol{\ell}, \boldsymbol{r}) & = \sum_{l = 1}^{l_\text{max}} \sum_{i=1}^{|\boldsymbol{\ell}|+|\boldsymbol{r}|} d_i^l(\boldsymbol{e}, \tilde{\boldsymbol{e}} \mid \boldsymbol{\ell}, \boldsymbol{r}) \\
d_i^l(\boldsymbol{e}, \tilde{\boldsymbol{e}} \mid \boldsymbol{\ell}, \boldsymbol{r}) & = \|\tilde{h}_i^l(\boldsymbol{\ell}, \boldsymbol{e}, \boldsymbol{r}) - \tilde{h}_i^l(\boldsymbol{\ell}, \tilde{\boldsymbol{e}}, \boldsymbol{r}) )  \|^2\,.
\end{align*}
That is, given an input sequence $[\boldsymbol{\ell};\boldsymbol{e};\boldsymbol{r}]$, $d_i^l(\boldsymbol{e}, \tilde{\boldsymbol{e}} \mid \boldsymbol{\ell}, \boldsymbol{r})$ measures the influence of replacing $\boldsymbol{e}$ with $\tilde{\boldsymbol{e}}$ on the contextualized representation of the $i$-th word in the $l$-th layer.

As $d(\boldsymbol{e}, \tilde{\boldsymbol{e}} \mid
\boldsymbol{\ell}, \boldsymbol{r})$ is differentiable with
respect to $\tilde{\boldsymbol{e}}$, we can 
use gradient-based optimization to estimate $\text{OTA}(w)$.
This idea resembles the approach of
\namecite{le2014distributed} to infer paragraph vectors for
sequences of arbitrary length.

With regards to the choice of contexts $\mathcal{C}$, we define two variants, both of which do not require any additional information: \textsc{static} and \textsc{random}. For the \textsc{static} variant, $\mathcal{C}$ consists of a single context 
\[
(\mathbf{t}_\ell,\mathbf{t}_r) = (\texttt{[CLS]},\ .\,\texttt{[SEP]})
\]
with $\texttt{[CLS]}$ and $\texttt{[SEP]}$ being BERT's classification and separation token, respectively. We use this particular context because in pretraining, BERT is exposed exclusively to sequences starting with \texttt{[CLS]} and ending with \texttt{[SEP]}.

As the meaning of a word can often better be understood by looking at its interaction with other words, we surmise that OTA works better when we provide variable contexts in which different words occur. For this reason, we also investigate the \textsc{random} variant. In this variant, each pair $(\mathbf{t}_\ell,\mathbf{t}_r) \in \mathcal{C}$ is of the form 
\[
(\mathbf{t}_\ell,\mathbf{t}_r) = (\texttt{[CLS]}\,t_\ell,\ t_r\,.\,\texttt{[SEP]})
\] 
where $t_\ell$ and $t_r$ are uniformly sampled tokens from
$\mathcal{T}$, under the constraint that each of them represent an actual word. 

\section{WordNet Language Model Probing}

In order to assess the ability of language models to understand words as a function of their frequency, we introduce the \emph{WordNet Language Model Probing} (\dataset) dataset.\footnote{The \dataset{} dataset is publicly available at \url{https://github.com/timoschick/am-for-bert}} 
This dataset consists of two parts:
\begin{itemize}
\item a set of triples $(k, r, T)$ where $k$ is a \emph{keyword}, $r$ is a \emph{relation} and $T$ is a set of \emph{target words};
\item a set of \emph{patterns} $P(r)$ for each relation $r$, where each pattern is a sequence of tokens that contains exactly one \emph{keyword placeholder} \keyword{} and one \emph{target placeholder} \mask.
\end{itemize}
The dataset contains four different kinds of relations: \textsc{antonym} (\textsc{ant}), \textsc{hypernym} (\textsc{hyp}), \textsc{cohyponym+} (\textsc{coh+}) and \textsc{corruption} (\textsc{cor}). Examples of dataset entries for all relations are shown in Table~\ref{ds-examples}; the set of patterns for each relation can be seen in Table~\ref{ds-patterns}.

\begin{table}
\begin{tabularx}{\linewidth}{llX}
\toprule
\textbf{Key} & \textbf{Rel.} & \textbf{Targets} \\
\midrule
new & \textsc{ant} & old \\
general & \textsc{ant} & specific \\
local & \textsc{ant} & global \\
\midrule
book & \textsc{hyp} & product, publication, \ldots \\
basketball & \textsc{hyp} & game, ball, sport, \ldots \\
lingonberry & \textsc{hyp} & fruit, bush, berry, \ldots \\
\midrule
samosa & \textsc{coh+} & pizza, sandwich, salad, \ldots \\
harmonium & \textsc{coh+} & brass, flute, sax, \ldots \\
immorality & \textsc{coh+} & crime, evil, sin, fraud, \ldots \\
\midrule
simluation & \textsc{cor} & simulation \\
chepmistry & \textsc{cor} & chemistry \\
pinacle & \textsc{cor} & pinnacle \\
\bottomrule
\end{tabularx}
\caption{Example entries from \dataset}
\label{ds-examples}
\end{table}

\begin{table}
\begin{tabularx}{\linewidth}{ll}
\toprule
\textsc{antonym} & \textsc{hypernym} \\
\midrule
\keyword{} is the opposite of \mask{} . & \keyword{} is a  \mask{} . \\
\keyword{} is not \mask{} . & a \keyword{} is a \mask{} . \\
someone who is \keyword{} is not \mask{} . & ``\keyword'' refers to a \mask{} . \\
something that is \keyword{} is not \mask{} . & \keyword{} is a kind of  \mask{} . \\
``\keyword{}'' is the opposite of ``\mask{}'' . & a \keyword{} is a kind of\,\mask{}\,.\\
\midrule
\textsc{corruption} & \textsc{cohyponym+} \\
\midrule
``\keyword{}'' is a misspelling of ``\mask{}'' . & \keyword{} and \mask{} .\\
``\keyword{}'' . did you mean ``\mask{}'' ? & ``\keyword{}'' and ``\mask{}'' .\\
\bottomrule
\end{tabularx}
\caption{Patterns for all relations of \dataset. The indefinite article ``a'' used in the \textsc{hyp} patterns is replaced with ``an'' as appropriate.}
\label{ds-patterns}
\end{table}

We split the dataset into a development and a test set. For each relation, we randomly select $10\%$ of all entries to be included in the development set; the remaining $90\%$ form the test set. We purposefully do not provide a training set as \dataset\  is meant to be used \emph{without} task-specific fine-tuning. We also define three subsets based on keyword counts in WWC: \dataset-\textsc{rare}, containing all words that occur less than 10 times, \dataset-\textsc{medium}, containing all words that occur 10 or more times, but less than 100 times, and \dataset-\textsc{frequent}, containing all remaining words. Statistics about the sizes of these subsets and the  mean number of target words per relation are listed in Table~\ref{ds-statistics}.

For creating \dataset{}, we use WordNet \cite{miller1995wordnet} to obtain triples $(k, r, T)$. To this end, we denote by $\mathcal{V}$ the vocabulary of all words that occur at least once in the Westbury Wikipedia Corpus (WWC) \cite{shaoul2010westbury} and match the regular expression $\texttt{[a}\text{--}\texttt{z.-]*}$. The set of all tokens in the BERT vocabulary is denoted by $\mathcal{T}$. For all triplets, we restrict the set of target words to single-token words from $\mathcal{T}$.
 This allows us to measure BERT's performance for each
 keyword $k$ without
the conflating influence of rare or multi-subword words on
 the target side.


\begin{table}
\newcolumntype{R}{>{\raggedleft\arraybackslash}X}
\begin{tabularx}{\linewidth}{lRRRRRR}
\toprule
& \multicolumn{3}{c}{\textbf{Subset Size}} & \multicolumn{3}{c}{\textbf{Mean Targets}}  \\
\cmidrule(lr){2-4}\cmidrule(lr){5-7}
\textbf{Rel.}& \multicolumn{1}{c}{\textsc{r}} & \multicolumn{1}{c}{\textsc{m}} & \multicolumn{1}{c}{\textsc{f}} & \multicolumn{1}{c}{\textsc{r}} & \multicolumn{1}{c}{\textsc{m}} & \multicolumn{1}{c}{\textsc{f}} \\
\midrule
\textsc{ant} & 41 & 59 & 266 & 1.0 & 1.0 & 1.0 \\
\textsc{hyp} & 1191 & 1785 & 4750 & 4.0 & 3.9 & 4.2 \\
\textsc{coh+} & 1960 & 2740 & 6126 & 26.0 & 26.0 & 25.0 \\
\textsc{cor} & 2880 & -- & -- & 1.0 & -- & -- \\
\bottomrule
\end{tabularx}
\caption{The number of entries and mean number of target words for the \textsc{rare} (\textsc{r}), \textsc{medium} (\textsc{m}), and \textsc{frequent} (\textsc{f}) subsets of \dataset}
\label{ds-statistics}
\end{table}

\sectionflex{Antonyms} For each adjective $w \in \mathcal{V}$, we collect all antonyms for its most frequent WordNet sense in a set $A$ and, if $A \cap \mathcal{T} \neq \emptyset$, add $(w,\textsc{antonym}, A \cap \mathcal{T})$ to the dataset.

\sectionflex{Hypernyms} For each noun $w \in \mathcal{V}$, let $H$ be the set of all hypernyms for its two most frequent senses. As direct hypernyms are sometimes highly specific (e.g., the hypernym of ``dog'' is ``canine''), we include all hypernyms whose path distance to $w$ is at most 3. To avoid the inclusion of very general terms such as ``object'' or ``unit'', we restrict $H$ to hypernyms that have a minimum depth of 6 in the WordNet hierarchy. If $|H \cap \mathcal{T}| \geq 3$, we add $(w, \textsc{hypernym}, H \cap \mathcal{T})$ to the dataset. However, if $|H \cap \mathcal{T}| > 20$, we keep only the $20$ most frequent target words.

\sectionflex{Cohyponyms+} For each noun $w \in \mathcal{V}$,
we compute its set of hypernyms $H$ as described above (but
with a maximum path distance of 2), and denote by $C$ the
union of all hyponyms for each hypernym in $H$ with a
maximum path distance of 4.\footnote{Cohyponyms are defined to
  have a common parent. Our more general definition (having
  a common ancestor) gives us a test that has
  more coverage than a restriction to cohyponyms in a strict
  sense would have. We call our generalization ``cohyponym+''.}
 Let $C' = (C \setminus \{w\}) \cap \mathcal{T}$. If ${|C'| \geq 10}$, we add the corresponding tuple $(w, \textsc{cohyponym+}, C')$ to the dataset. If $|C'| > 50$, we keep only the $50$ most frequent target words.

\sectionflex{Corruptions}
We include this relation to investigate a model's ability to deal with corruptions of the input that may, for example, be the result of typing errors or errors in optical character recognition.
To obtain corrupted words, we take frequent words
from $\mathcal{V} \cap \mathcal{T}$ and randomly apply
corruptions similar to the ones used by
\namecite{hill2016learning} and
\namecite{lee2018deterministic}, but we apply them on the
character level. Specifically, given a word $w = c_1 \ldots c_n$, we create a corrupted version $\tilde{w}$ by either (i) inserting a random character $c$ after a random position $i \in [0, n]$, (ii) removing a character at a random position $i \in [1,n]$ or (iii) switching the characters $c_i$ and $c_{i+1}$ for a random position $i \in [1, n-1]$. We then add $(\tilde{w}, \textsc{corruption}, w)$ to the dataset.

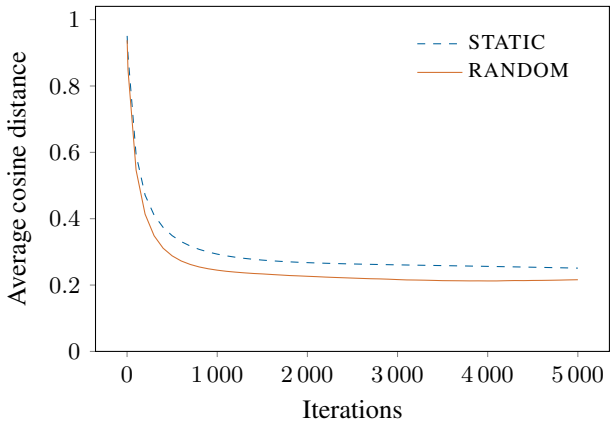
\begin{figure}
\centering
\begin{tikzpicture}
\begin{axis}[
	xlabel={Iterations},
		ylabel={Average cosine distance},
    ymin = 0,
    ymax = 1.04,
    xmin = -350,
    xmax = 5350,
    xtick pos=left,
    ytick pos=left,
        ylabel near ticks,
        xlabel near ticks,
    tick align=outside,
          major tick length=0.075cm,
    width = \linewidth,
    height = 0.27\textheight,
    x tick label style={/pgf/number format/1000 sep=\,},
    legend style={at={(0.95,0.95)},anchor=north east, draw=none},
    legend cell align=left,
    legend columns=1,
    tick label style={font=\footnotesize}
]

\addplot[color=col1, dashed, mark options={solid}] coordinates {
(1,0.9510249543357931)
(10,0.8947415728124324)
(100,0.5930100318118929)
(200,0.47114215180277824)
(300,0.4107090431600809)
(400,0.3738197379410267)
(500,0.34890741735696795)
(600,0.33105666372179987)
(700,0.3177554743587971)
(800,0.3075638880431652)
(900,0.2995853697359562)
(1000,0.2932390650808811)
(1100,0.2881197871565819)
(1200,0.2839412495493889)
(1300,0.28049922797083854)
(1400,0.277634765714407)
(1500,0.27523286816477777)
(1600,0.2732065429389477)
(1700,0.27147555896639824)
(1800,0.26999268132448195)
(1900,0.2687071707844734)
(2000,0.26759131741523745)
(2100,0.2666080770492554)
(2200,0.26573490029573443)
(2300,0.264953654140234)
(2400,0.26424363803863526)
(2500,0.263594192802906)
(2600,0.2630032795369625)
(2700,0.2624431154727936)
(2800,0.2619158705770969)
(2900,0.2614110948443413)
(3000,0.260929039478302)
(3100,0.26046122395992277)
(3200,0.25999704509973526)
(3300,0.2595439055263996)
(3400,0.2590929105579853)
(3500,0.25863046661019323)
(3600,0.2581791011989117)
(3700,0.25771632918715476)
(3800,0.2572498868703842)
(3900,0.2567750377655029)
(4000,0.2562956999838352)
(4100,0.255798135638237)
(4200,0.2553053259849548)
(4300,0.25479394191503524)
(4400,0.25427893418073655)
(4500,0.2537556767761707)
(4600,0.2532237367033958)
(4700,0.25267999538779257)
(4800,0.25212747222185133)
(4900,0.251574894785881)
(5000,0.25101661124825475)
};
\addlegendentry{\textsc{static}};

\addplot[color=col2] coordinates {
(1,0.9377327941135154)
(10,0.8500339128412306)
(100,0.5464558956921101)
(200,0.4137031869292259)
(300,0.348594661206007)
(400,0.31102063575387)
(500,0.28803888446092607)
(600,0.2726473034918308)
(700,0.26219887217879295)
(800,0.2545870494544506)
(900,0.24909069532155992)
(1000,0.2448794022500515)
(1100,0.24151456826925277)
(1200,0.23901816335320472)
(1300,0.23682504642009736)
(1400,0.23493026611208917)
(1500,0.2336080761551857)
(1600,0.2319061544239521)
(1700,0.23035698908567429)
(1800,0.22871226304769515)
(1900,0.22768330782651902)
(2000,0.22645808586478233)
(2100,0.22535821679234505)
(2200,0.22404028898477554)
(2300,0.22296742549538612)
(2400,0.2220181883573532)
(2500,0.22102995923161506)
(2600,0.2200446575284004)
(2700,0.21910717257857323)
(2800,0.21857947105169295)
(2900,0.21769134441018104)
(3000,0.21652068603038788)
(3100,0.21548574364185333)
(3200,0.21515098854899406)
(3300,0.2146763531565666)
(3400,0.21403382313251496)
(3500,0.21330079287290574)
(3600,0.21310225489735604)
(3700,0.21288260811567306)
(3800,0.21255865600705146)
(3900,0.21267574855685234)
(4000,0.21241310769319535)
(4100,0.21245131957530974)
(4200,0.21310457998514176)
(4300,0.213595038741827)
(4400,0.2134302940070629)
(4500,0.21389037045836448)
(4600,0.21407312068343162)
(4700,0.21445456898212434)
(4800,0.21494768393039704)
(4900,0.21557593327760696)
(5000,0.21604415094852447)
};
\addlegendentry{\textsc{random}};

\end{axis}
\end{tikzpicture}
\caption{Performance of OTA on 1000 randomly selected one-token words}
\label{ota-hyperparams}
\end{figure}

\section{Experiments}

For our evaluation of BERT on \dataset{}, we use the Transformers library of \namecite{wolf2019transformers}. Our implementation of OTA is based on PyTorch \cite{paszke2017automatic}.\footnote{Our implementation of OTA is publicly available at \url{https://github.com/timoschick/one-token-approximation}}
For all of our experiments involving AM, we use the original implementation of \namecite{schick2019attentive}. As \dataset{} is based on WordNet, all of our experiments are confined to the English language.

\subsection{One-Token Approximation}

We first compare the \textsc{static} and
\textsc{random} context variants of OTA and determine the
optimal number of training iterations. To this end, we form a development set by
randomly selecting 1000 one-token words from the BERT
vocabulary. For each word $w$ in this set, we measure the
quality of its approximation $\text{OTA}(w)$ by comparing it
to its BERT embedding $e(w)$, using cosine distance. We
  initialize the OTA vector of each word as a zero vector
  and optimize it using Adam \cite{kingma2014adam} with an initial learning rate of $10^{-3}$. For
both context variants, we search for the ideal number of
iterations in the range $\{100\cdot i \mid 1 \leq i \leq 50
\}$. 

Results can be seen in Figure~\ref{ota-hyperparams}. While for both variants, the average cosine distance between BERT's embeddings and their OTA equivalents is relatively high in the beginning -- which is simply due to the fact that all OTA embeddings are initialized randomly -- after only a few iterations
\textsc{random}
consistently
outperforms
\textsc{static}.\footnote{The difference between the best results achieved using \textsc{random} and \textsc{static} is statistically significant in a two-sided binomial test ($p < 0.05$).} For the \textsc{random} variant, the
average cosine distance reaches its minimum at 4000
iterations. We therefore  use \textsc{random} contexts
with 4000 iterations in our following experiments.

\subsection{Evaluation on \dataset}

To measure the performance of a language model on \dataset,
we proceed as follows. Let $x = (k,r,T)$ be a dataset entry,
$w \in T$ a target word, $p \in P(r)$  a pattern and
$p[k]$ the same pattern where the keyword placeholder
\keyword{} is replaced by $k$. Furthermore, let $(a_1,
\ldots, a_n)$ be the model's responses (sorted in descending
order by their probability) when it is asked to predict a
replacement word for the target placeholder in $p[k]$. Then
there is some $j$ such that $a_j = w$. We denote with
\begin{align*}
\text{rank}(p[k],w) & = j \\ 
\text{precision}_{i}(p[k],T) & = \frac{|\{ a_1, \ldots, a_i \} \cap T|}{i}
\end{align*}
the \emph{rank of $w$} and \emph{precision at $i$} when the model is queried with $p[k]$.\footnote{We only look at the first $100$ system responses and set $\text{rank}(p[k],w) = \infty$ if $w \notin \{ a_1, \ldots, a_{100} \}$.} We may then define:
\begin{align*}
\text{rank}(x) & = \min_{p \in P(r)} \min_{w \in T} \text{rank}(p[k],w) \\
\text{precision}_i(x) & = \max_{p \in P(r)} \text{precision}_{i}(p[k],T)
\end{align*}
That is, for each triplet $x$, we compute the \emph{best} rank and precision that can be achieved using any pattern. We do so because our interest is not in testing the model's ability to understand a given pattern, but its ability to understand a given word: by letting the model choose the best pattern for each word, we minimize the probability that its response is of poor quality simply because it did not understand a given pattern.

We evaluate the uncased version of \bertbase{} \cite{devlin2018bert}
on  \dataset{} to
get an impression of (i) the model's general ability to
understand the presented phrases and (ii) the difference in
performance for rare and frequent words. 
To investigate how well OTA does at obtaining single embeddings for multi-token words, we also try a variant of BERT where all multi-token keywords are replaced with their one-token approximations. Furthermore, we compare OTA against the following baseline strategies for obtaining single embeddings for multi-token words $w = t_1, \ldots, t_n$:

\begin{itemize}
\item \textsc{first}: We use the embedding of the first token, $e(t_1)$.
\item \textsc{last}: We use the embedding of the last token, $e(t_n)$.
\item \textsc{avg}: We use the average over the embeddings of all tokens, $\frac{1}{n}\sum_{i=1}^n e(t_i)$.
\end{itemize}
We choose these particular baselines because they are natural choices for obtaining a word embedding from a sequence of subword embeddings without any advanced computation.

\begin{figure}
\centering
\begin{tikzpicture}
\begin{axis}[
	cycle list name=color,
	xlabel={\dataset\ Subset},
		ylabel={MRR},
    ymin = 0,
    ymax = 0.42,
    xmin = -0.2,
    xmax = 2.2,
    xtick pos=left,
    ytick pos=left,
        ylabel near ticks,
        xlabel near ticks,
    tick align=outside,
          major tick length=0.075cm,
    width = \linewidth,
    height = 0.27\textheight,
    xtick={0,1,2},
    xticklabels={\textsc{rare}, \textsc{medium}, \textsc{freq}.},
    ytick={0, 0.1, 0.2, 0.3, 0.4},
    legend style={at={(0.05,0.95)},anchor=north west, draw=none, fill=none},
    legend cell align=left,
    legend columns=1,
    tick label style={font=\footnotesize}
]

\addplot[mark=*, mark options={solid}, color=col1]  coordinates {
(0,0.112)
(1,0.229)
(2,0.350)
};
\addlegendentry{\scriptsize \bertbase};

\addplot[dashed, mark=triangle*, mark options={solid}, color=col2]  coordinates {
(0,0.107)
(1,0.216)
(2,0.332)
};
\addlegendentry{\scriptsize{OTA}};



\addplot[dashed, mark=*, mark options={solid, fill opacity=0}, color=col3]  coordinates {
(0,0.063)
(1,0.071)
(2,0.190)
};
\addlegendentry{\scriptsize{\textsc{first}}};

\addplot[dotted, mark=square, mark options={solid}, color=col4]  coordinates {
(0,0.033)
(1,0.082)
(2,0.196)
};
\addlegendentry{\scriptsize{\textsc{last}}};

\addplot[dashdotted, mark=triangle, mark options={solid}, color=col5]  coordinates {
(0,0.033)
(1,0.073)
(2,0.201)
};
\addlegendentry{\scriptsize{\textsc{avg}}};

\end{axis}
\end{tikzpicture}
\caption{Mean reciprocal rank on \dataset\ dev+test for \bertbase{}, OTA and various baselines}
\label{bert-psr}
\end{figure}
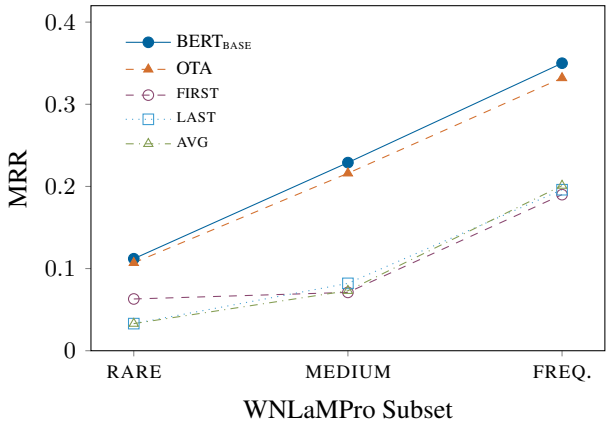

The mean reciprocal rank (MRR) over \dataset\ can be seen in Figure~\ref{bert-psr} for \bertbase{}, OTA and all baselines. We can see that for all models, the score depends heavily on the word frequency.
Notably, OTA performs much better than all of the above
baselines, regardless of word frequency. Furthermore, the
difference in performance between OTA's single embeddings
and BERT's original, multi-token embeddings is only
marginal, allowing us to conclude that OTA is indeed able to
infer single-token embeddings of decent quality for
multi-token words.

\begin{figure}
\centering
\begin{tikzpicture}
\begin{axis}[
	width=0.885\linewidth,
	height=0.27\textheight,
    view={0}{90},   
    xlabel={Word count},
    ylabel={Rank},
    colorbar,
    colormap = {whiteblack}{color(0cm)  = (white);color(1cm) = (black)},
    colorbar style={
        yticklabel style={
            /pgf/number format/.cd,
            fixed,
            precision=1,
            fixed zerofill,
        },
        width=0.2cm,
    },
    %
    enlargelimits=false,
    axis on top,
    point meta min=0.06,
    point meta max=0.34,
    %
    ymin=-0.5,
    ymax=10.5,
    xmin=-0.5,
    xmax=14.5,
    xtick={0,2,4,6,8,10,12,14},
    xticklabels={$2^0$,$2^2$,$2^4$,$2^6$, $2^8$, $2^{10}$, $2^{12}$, $\infty$},
    ytick={0,1,2,3,4,5,6,7,8,9,10},
    yticklabels={$1$, $2$, $4$, $8$, $16$, $32$, $64$, $128$, $256$, $512$, $\infty$}, 
    xtick pos=left,
    ytick pos=left,
    ylabel near ticks,
    xlabel near ticks,
    tick align=outside,
    major tick length=0.075cm,
    tick label style={font=\footnotesize}
]

\addplot[matrix plot*,point meta=explicit] table [x=xval, y=yval, meta=z, col sep=comma] {ranks.csv};

\end{axis}
\end{tikzpicture}
\caption{Performance of \bertbase{} for
	the \textsc{cohyponym+} subset of \dataset. Each
	cell $(i,j)$ of the  heat map is shaded based on the
	percentage of all dataset entries with keyword
	counts
(``Word count'')
in the range $(2^{j-1}, 2^j]$ whose rank (``Rank'') is in the range $(2^{i-1}, 2^i]$. The values in each column add up to one.}
\label{heatmap}
\end{figure}
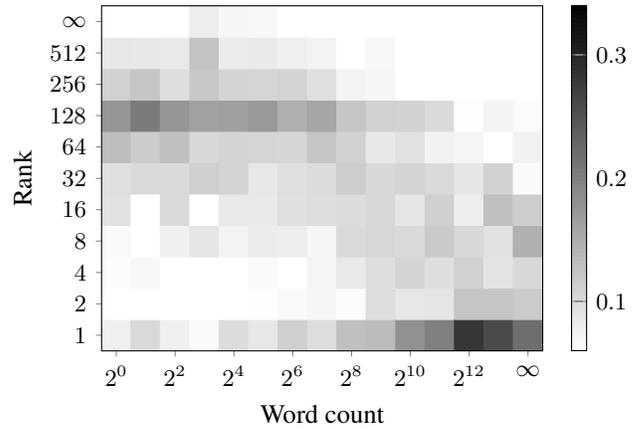

Of course, OTA by itself  does not improve the embedding
quality compared to using BERT as is
-- and we never apply OTA to words that have single-token
BERT representations in the following experiments. The
purpose of OTA is to allow us to train our attentive mimicking model for BERT:
OTA provides us with the single-token embeddings that we require to train AM.
  
The general trend that the understanding of a word increases
with its frequency becomes even more obvious when looking at
Figure~\ref{heatmap}, where the distribution of ranks for
the \textsc{cohyponym+} subset of \dataset\ is shown as a
function of WWC word counts. The distribution of ranks is computed independently for each interval of word counts considered. That is, the values in each column are normalized so that they add up to one. This was done to prevent the diagram from being distorted because certain word count intervals contain more words than others.
As can be seen, for words that occur at most 256 ($2^8$) times in  WWC, the most probable
rank interval is $[64, 128)$. With more observations, BERT's
  understanding of words drastically improves: more than
  50\% of all words with more than 256 ($2^8$) observations achieve a rank of at most 16.

\begin{table}
\newcolumntype{Y}{>{\centering\arraybackslash}X}
\begin{tabularx}{\linewidth}{XYY}
\toprule
& \multicolumn{2}{c}{\textbf{MRR}} \\
\cmidrule(lr){2-3}
\textbf{Model} & \multicolumn{1}{c}{5 Epochs} & \multicolumn{1}{c}{10 Epochs} \\
\midrule
AM & 0.258 & 0.253 \\
AM+\textsc{context} & \textbf{0.262} & \textbf{0.276} \\
\midrule
AM $-$ OTA & 0.219 & 0.220 \\
AM $-$ form & 0.138 & 0.133 \\
AM $-$ context & 0.227 & 0.225 \\
\bottomrule
\end{tabularx}
\caption{Results on \dataset\ dev for various configurations of AM trained on embeddings from and integrated into \bertbase}
\label{am-hyperparams}
\end{table}

\begin{table*}
\small
\newcolumntype{R}{>{\raggedleft\arraybackslash}X}
\begin{tabularx}{\linewidth}{llRRRRRRRRR}
\toprule
&& \multicolumn{3}{c}{\textsc{rare}} & \multicolumn{3}{c}{\textsc{medium}} & \multicolumn{3}{c}{\textsc{frequent}} \\
\cmidrule(lr){3-5}\cmidrule(lr){6-8}\cmidrule(lr){9-11}
\textbf{Set}&\textbf{Model} & MRR & P@3 & P@10 & MRR & P@3 & P@10 & MRR & P@3 & P@10 \\
\midrule
\multirow{4}{*}{\textsc{ant}} & \bertbase{} & 0.149 & 0.065 & 0.025 & 0.089 & 0.044 & 0.021 & 0.390 & 0.170 & 0.061 \\
 & \bertbase{} + AM & \underline{\textbf{0.449}} & \underline{\textbf{0.167}} & \underline{\textbf{0.075}} & \underline{\textbf{0.511}} & \underline{\textbf{0.176}} & \underline{\textbf{0.064}} & \underline{\textbf{0.482}} & \underline{\textbf{0.195}} & \underline{\textbf{0.074}} \\
 \cdashlinelr{2-11}
 & \bertlarge{} & 0.234 & 0.083 & 0.044 & 0.218 & 0.088 & 0.036 & 0.541 & 0.209 & 0.081 \\
 & \bertlarge{} + AM & \underline{\textbf{0.529}} & \underline{\textbf{0.194}} & \underline{\textbf{0.075}} & \underline{\textbf{0.558}} & \underline{\textbf{0.195}} & \underline{\textbf{0.068}} & \textbf{0.570} & \underline{\textbf{0.228}} & \underline{\textbf{0.088}} \\
\midrule
\multirow{4}{*}{\textsc{hyp}} & \bertbase{} & 0.276 & 0.122 & 0.066 & 0.327 & 0.151 & 0.077 & \underline{\textbf{0.416}} & \underline{\textbf{0.204}} & \underline{\textbf{0.109}} \\
 & \bertbase{} + AM & \underline{\textbf{0.300}} & \underline{\textbf{0.135}} & \underline{\textbf{0.074}} & \textbf{0.343} & \textbf{0.158} & \underline{\textbf{0.081}} & 0.377 & 0.181 & 0.096 \\
  \cdashlinelr{2-11}
 & \bertlarge{} & 0.284 & 0.128 & 0.065 & \underline{\textbf{0.350}} & \underline{\textbf{0.169}} & \underline{\textbf{0.086}} & \underline{\textbf{0.462}} & \underline{\textbf{0.226}} & \underline{\textbf{0.117}} \\
 & \bertlarge{} + AM & \underline{\textbf{0.299}} & \textbf{0.137} & \underline{\textbf{0.074}} & 0.323 & 0.149 & 0.079 & 0.401 & 0.193 & 0.101 \\
\midrule
\multirow{4}{*}{\textsc{coh+}} & \bertbase{} & 0.147 & 0.065 & 0.054 & 0.177 & 0.089 & 0.070 & \underline{\textbf{0.294}} & \underline{\textbf{0.150}} & \underline{\textbf{0.116}} \\
 & \bertbase{} + AM & \underline{\textbf{0.213}} & \underline{\textbf{0.106}} & \underline{\textbf{0.082}} & \underline{\textbf{0.213}} & \underline{\textbf{0.110}} & \underline{\textbf{0.090}} & 0.262 & 0.136 & 0.108 \\
  \cdashlinelr{2-11}
 & \bertlarge{} & 0.174 & 0.085 & 0.067 & 0.210 & \textbf{0.109} & \textbf{0.091} & \underline{\textbf{0.337}} & \underline{\textbf{0.183}} & \underline{\textbf{0.143}} \\
 & \bertlarge{} + AM & \underline{\textbf{0.227}} & \underline{\textbf{0.110}} & \underline{\textbf{0.087}} & \textbf{0.216} & 0.106 & 0.089 & 0.292 & 0.153 & 0.121 \\
\midrule
\multirow{4}{*}{\textsc{cor}} & \bertbase{} & 0.020 & 0.007 & 0.004 & -- & -- & -- & -- & -- & -- \\
 & \bertbase{} + AM & \underline{\textbf{0.254}} & \underline{\textbf{0.095}} & \underline{\textbf{0.038}} & -- & -- & -- & -- & -- & -- \\
  \cdashlinelr{2-11}
 & \bertlarge{} & 0.062 & 0.022 & 0.012 & -- & -- & -- & -- & -- & -- \\
 & \bertlarge{} + AM & \underline{\textbf{0.261}} & \underline{\textbf{0.095}} & \underline{\textbf{0.038}} & -- & -- & -- & -- & -- & -- \\
\bottomrule
\end{tabularx}

\caption{Performance of BERT with and without AM
  for \dataset\ test, subdivided by relation and keyword count. Underlined numbers indicate a significant difference between BERT and BERT+AM in a two-sided binomial test ($p < 0.05$).}
\label{main-results}
\end{table*}

\subsection{Attentive Mimicking}

We train two variants of Attentive Mimicking: the default
configuration of \namecite{schick2019attentive} and the
AM+\textsc{context} configuration (\secref{amcontext}) that
puts more emphasis on contexts. To decide which method to
apply and to determine the optimal number of training
epochs, we use \dataset\ dev. As evaluating AM on \dataset\ is a time-consuming operation, the only values we try are $5$ and $10$ epochs; furthermore, we perform hyperparameter optimization only on \bertbase{}. 
To understand the influence of one-token approximation on
the performance of AM, in addition to the two configurations
described above -- both of which make use of OTA -- we also
try a variant without OTA, where the training set contains
only one-token words. To see whether we actually need both
form and context information, we additionally investigate
the influence of dropping either the context or form parts of AM.

As proposed by \namecite{schick2018learning}, we train AM on
all words that occur at least $100$ times in WWC; for each
word that is represented by multiple tokens in the BERT
vocabulary, we use its OTA as a target vector to be
mimicked. Importantly, we train AM on contexts from WWC
(containing slightly fewer than $10^9$ words), whereas the
original BERT model was trained on the concatenation of
BooksCorpus \cite{zhu2015aligning} (containing $0.8 \cdot
10^9$ words) and a larger version of Wikipedia (containing
$2.5 \cdot 10^9$ words). 
Each
occurrence of a word can contribute to obtaining a high-quality
representation,
especially for rare words. Therefore, 
BERT has a clear advantage
over our proposed method due to its larger training corpus.

Table~\ref{am-hyperparams} shows results for all model
variants on \dataset\ dev. We can see that OTA is indeed helpful for training the model, substantially improving its score. Results for the model variants using only form or context are in line with the findings of \namecite{schick2018learning}:  it is essential for good performance to use both form and context. Furthermore, AM+\textsc{context} improves upon the default configuration of AM and training it for $10$ epochs performs better than $5$ epochs. 
Based on these findings, we only apply AM+\textsc{context} trained for 10 epochs using OTA on \dataset{} test.

For both the base and large configurations of BERT, Table~\ref{main-results} compares
BERT's performance with and without
AM on \dataset{};
MRR as well as precision at 3 and 10 are shown for each
relation and frequency.  AM substantially improves the score
for rare words, both for \bertbase{} and for
\bertlarge{}. The difference between BERT with and without
AM is significant according to a two-sided binomial test ($p
< 0.05$).
This demonstrates that AM  helps BERT  get a better understanding of rare words. The benefit of applying AM for medium frequency words depends largely on the model being used: for \bertlarge{}, using AM only brings a consistent improvement for the \textsc{antonym} relation, whereas for \bertbase{}, using AM is always helpful. The fact that BERT performs better than AM for frequent words is not surprising, considering that our model both has less capacity and was trained on considerably less data. However, the strong results for rare and -- in some cases -- medium frequency words suggest that to obtain the best of both worlds, one can simply replace BERT's embeddings for rare words using AM while keeping its original embeddings for frequent words. As AM is trained using mimicking as an objective, embeddings induced by AM are well aligned with the embedding space it was trained on. Thus, BERT's original embeddings and AM-based embeddings can seamlessly be employed together.

To better understand for what kinds of words adding AM to
BERT is especially helpful, we finally analyze the
predictions of BERT with and without AM for a few selected
words (Table~\ref{qualitative-examples}). As exemplified by
these examples, the inability of BERT to understand rare
words is often due to the tokenization algorithm splitting
words in a suboptimal way
(``una$\cdot$cc$\cdot$ess$\cdot$ible'' and
``un$\cdot$ic$\cdot$y$\cdot$cle'' instead of
``un$\cdot$access$\cdot$ible'' and ``uni$\cdot$cycle''). As
AM uses overlapping $n$-grams to represent a word's surface
form and thus does not need to choose a single tokenization,
it does not suffer from that problem. While BERT's tokenization problem could potentially also be addressed by replacing WordPiece with a morphology-aware tokenization algorithm, other words -- such
  as ``salsify'' -- simply cannot be decomposed into smaller
  meaningful units. BERT also struggles
  with spelling errors (e.g., ``resigntaion'') and rare
  spellings (e.g., ``bulghur'', ``kidnaper'').

\begin{table}
\small
\begin{tabularx}{\linewidth}{lX}
\toprule
\textbf{Query}:& something that is \emph{una$\cdot$cc$\cdot$ess$\cdot$ible} is not \mask{} . \\
\textbf{BERT}:& possible, impossible, true, allowed \\ 
\textbf{BERT+AM}:& accessible, allowed, possible, available \\
\midrule
\textbf{Query}:& \emph{un$\cdot$ic$\cdot$y$\cdot$cle} and \mask{} . \\
\textbf{BERT}:& bridge, body, base, chain \\ 
\textbf{BERT+AM}:& bicycle, pedestrian, walking, pedestrians \\
\midrule
\textbf{Query}:& a \emph{sal$\cdot$si$\cdot$fy} is a \mask{} . \\
\textbf{BERT}:& cocktail, toilet, noun, boat \\ 
\textbf{BERT+AM}:& shrub, flower, plant, noun \\
\midrule
\textbf{Query}:& `` \emph{resign$\cdot$tai$\cdot$on} '' is a misspelling of `` \mask{} '' .\\
\textbf{BERT}:& king, john, son, death \\
\textbf{BERT+AM}:& resignation, resign, resigned, resigning \\
\bottomrule
\end{tabularx}
\caption{Example queries from \dataset{} and most probable outputs of \bertbase{} and BERT+AM. The tokenization of keywords used by BERT is indicated by $\cdot$ characters.}
\label{qualitative-examples}
\end{table}

\section{Conclusion}

We have introduced \dataset, a new dataset that allows us to
explicitly investigate the ability of
language models to understand rare words. Using this
dataset, we have shown that BERT struggles with words if
they are too rare. To address this problem, we proposed to
apply Attentive Mimicking (AM). For AM to work, we
introduced
one-token approximation
(OTA), an effective method to obtain ``single-token'' embeddings for
multi-token words. Using this method, we showed that AM is
able to substantially improve BERT's understanding of rare
words.

Future work might investigate whether more complex
architectures than AM can bring further benefit to deep
language models; it would also be interesting to see whether
training AM on a larger corpus -- such as the one used for
training BERT by \namecite{devlin2018bert} -- is
beneficial. Furthermore, it would be interesting to see the
impact of integrating AM on downstream tasks.

\mbox{\ }

\section*{Acknowledgments}
This work was funded by the European Research Council (ERC \#740516).
We would like to thank the anonymous reviewers
for their helpful comments and their willingness to engage
with our author response.

\bibliography{AAAI-SchickT.4255.bibliography}
\bibliographystyle{aaai}
\end{document}